\documentclass[letterpaper, 10 pt, conference]{ieeeconf}  

\IEEEoverridecommandlockouts                  

\usepackage{graphics} 
\usepackage{epsfig} 
\usepackage{mathptmx} 
\usepackage{times} 
\usepackage{amsmath} 
\usepackage{amssymb}  
\usepackage{multirow}
\usepackage{makecell}
\usepackage{hyperref}
\usepackage{booktabs}

\title{\LARGE \bf
CVVLSNet: Vehicle Location and Speed Estimation Using Partial Connected Vehicle Trajectory Data
}

\author{Jiachen YE${^1}$, Dingyu WANG${^2}$, Shaocheng JIA${^{3, *}},$ Xin PEI${^{4, *}}$, Zi YANG${^5}$, Yi ZHANG${^2}$, and S.C. WONG${^3}$
\thanks{This work has been accepted by The 27th IEEE International Conference on Intelligent Transportation Systems (IEEE ITSC 2024).}
\thanks{$^{1}$Jiachen YE is with the Department of Civil Engineering, Tsinghua University, Beijing, China.}%
\thanks{$^{2}$Dingyu WANG and Yi ZHANG are with the Department of Automation, Beijing National Research Center for Information Science and Technology, Tsinghua University, Beijing, China.}%
\thanks{$^{3}$Shaocheng JIA and S.C. WONG are with the Department of Civil Engineering, The University of Hong Kong, Hong Kong, China.}%
\thanks{$^{4}$Xin PEI is with the Department of Automation, Beijing National Research Center for Information Science and Technology, Tsinghua University, Beijing, China and Shanghai Artificial Intelligence Laboratory, Shanghai, China. }%
\thanks{$^{5}$Zi YANG is with the School of Automation, Nanjing University of Science and Technology, Nanjing, China and Shanghai Artificial Intelligence Laboratory, Shanghai, China. }%
\thanks{$^{*}$Corresponding authors: Shaocheng JIA and Xin PEI; E-mail:  \href{shaocjia@connect.hku.hk}{shaocjia@connect.hku.hk}, \href{peixin@mail.tsinghua.edu.cn}{peixin@mail.tsinghua.edu.cn}.}%
}

\begin{document}

\maketitle
\thispagestyle{empty}
\pagestyle{empty}

\begin{abstract}

Real-time estimation of vehicle locations and speeds is crucial for developing many beneficial transportation applications in traffic management and control, e.g., adaptive signal control. Recent advances in communication technologies facilitate the emergence of connected vehicles (CVs), which can share traffic information with nearby CVs or infrastructures. At the early stage of connectivity, only a portion of vehicles are CVs. The locations and speeds for those non-CVs (NCs) are not accessible and must be estimated to obtain the full traffic information. To address the above problem, this paper proposes a novel CV-based Vehicle Location and Speed estimation network, CVVLSNet, to simultaneously estimate the vehicle locations and speeds exclusively using partial CV trajectory data. A road cell occupancy (RCO) method is first proposed to represent the variable vehicle state information. Spatiotemporal interactions can be integrated by simply fusing the RCO representations. Then, CVVLSNet, taking the Coding-RAte TransformEr (CRATE) network as a backbone, is introduced to estimate the vehicle locations and speeds. Moreover, physical vehicle size constraints are also considered in loss functions. Extensive experiments indicate that the proposed method significantly outperformed the existing method under various CV penetration rates, signal timings, and volume-to-capacity ratios. 

\end{abstract}

\section{INTRODUCTION}

Vehicle states, including the locations and speeds of all vehicles on the road, can reflect traffic conditions in a straightforward and detailed manner. Real-time estimation of vehicle locations and speeds is vital to various transportation applications, such as traffic signal control \cite{1}, \cite{2}, congestion prediction \cite{3}, \cite{4}, etc. Traditional methods rely on various forms of external information, such as roadside cameras \cite{5} and checkpoint data \cite{6}, to obtain vehicle states. These methods are constrained by the involvement of roadside infrastructures. Given the fixed positions of infrastructures, it is challenging to monitor all vehicles on the road. 

The CV technology provides an effective approach for vehicle state estimation \cite{7}. CVs can share traffic information with each other, including time, location, speed, etc. \cite{8}. These messages can benefit various transport applications, such as adaptive signal control. However, the transition period to full CV deployment is prolonged \cite{9}, \cite{10}. This means that a mixed traffic environment is inevitable, where both CVs and NCs exist in the network. Therefore, the locations and speeds of those NCs must be estimated using the accessible CV data to recover the full vehicle information on the road. To the best of our knowledge, no existing work applies advanced deep neural networks to estimating vehicle locations and speeds based on CV data. 

To address the above problem, this study proposes a CVbased Vehicle Location and Speed network, CVVLSNet, to estimate vehicle locations and speeds. The proposed method is visual-detector-free and relies solely on the limited CV trajectory data. Thus, it is economical compared with the visual-detector-based methods. Moreover, the method is robust to low visibility. Experiments show that CVVLSNet significantly outperformed the existing methods under various CV penetration rates and volume-to-capacity (V/C) ratios. Furthermore, the CRATE network used in CVVLSNet ensures mathematical interpretability, so that the computational logic can be clearly reproduced when giving parameters. 

The rest of the paper is organized as follows. Section II reviews the related works. Section III defines the problem. Section IV introduces the method. Section V conducts experiments. Section VI concludes the paper. Section VII discusses the limitations and future work. 

\section{RELATED WORKS}

The main methods for estimating vehicle states are based on various visual sensors \cite{5}, \cite{11}, \cite{12}. In particular, roadside cameras are the most widely used in estimating vehicle locations and speeds. For example, depth estimation methods can be used to calculate the distance between the vehicles and the camera, thereby determining the location of the vehicle \cite{13}. Some specific features, such as the shadows of the cars, were used to assist in vehicle distance measurement \cite{14}. Meanwhile, the displacement of the vehicle can be obtained through feature matching between adjacent frames so as to estimate the speed of the vehicle \cite{15}. The features used for vehicle matching include the bounding boxes of object detection \cite{16} and visual features of vehicles \cite{17}. In addition, semantic segmentation algorithms were also used to improve feature matching \cite{18}. Nevertheless, the involvement of roadside cameras limits their universal use.  

Comparatively, CV trajectory-based approach is more flexible and promising. To the best of our knowledge, the EVLS algorithm is the only existing method of estimating unequipped vehicles’ locations and speeds using CV trajectory data \cite{1}. In EVLS algorithm, the CV penetration rate and car-following model are assumed to be known. However, both remain unknown in reality, resulting in unsatisfactory performance in many situations. 

With the rapid development of deep learning, deep neural networks have been applied to various transportation applications \cite{19}, \cite{20}, \cite{21}, \cite{22}, \cite{23}, \cite{24}, \cite{25}, \cite{26}, \cite{27}, \cite{28}, \cite{29}. However, most deep learning networks are black box and lack interpretability. In contrast, the white-box deep learning framework \cite{30}, \cite{31}, \cite{32} has better mathematical interpretability for downstream tasks. Thus, it has great potential to be used in transportation applications, especially in safety-related tasks. 

Taking all pros and cons of the above methods into account, this study employs an interpretable white-box deep learning model as backbone to estimate vehicle locations and speeds in mixed traffic environment. The proposed method does not rely on any detector data and can accurately estimate the vehicle locations and speeds in diverse scenarios. 

\section{PROBLEM STATEMENT}

Consider a link with $l$ lanes and the length of $d$. The sets of CV states and full vehicle states in the link are denoted as 
\begin{align}
S_{X}(t)&=\{s_{c_1},s_{c_2},s_{c_3},\cdots \}, \nonumber \\
    S_{Y}(t)&=\{s_{c_1},s_{c_2},s_{c_3},\cdots;\  s_{n_1},s_{n_2},s_{n_3},\cdots\},
\end{align}
\noindent where $S_{X}$ is the set of CV data; $S_{Y}$ is the set of full vehicle data; $t$ represents the time of interest; $s_i$ is a tuple and includes the location and speed of the vehicle $i$; $c_j$ represents $j^{th}$ CV; $n_j$ represents $j^{th}$ NC. 
\begin{figure}[thpb]
      \centering
      \includegraphics[width=0.85\linewidth]{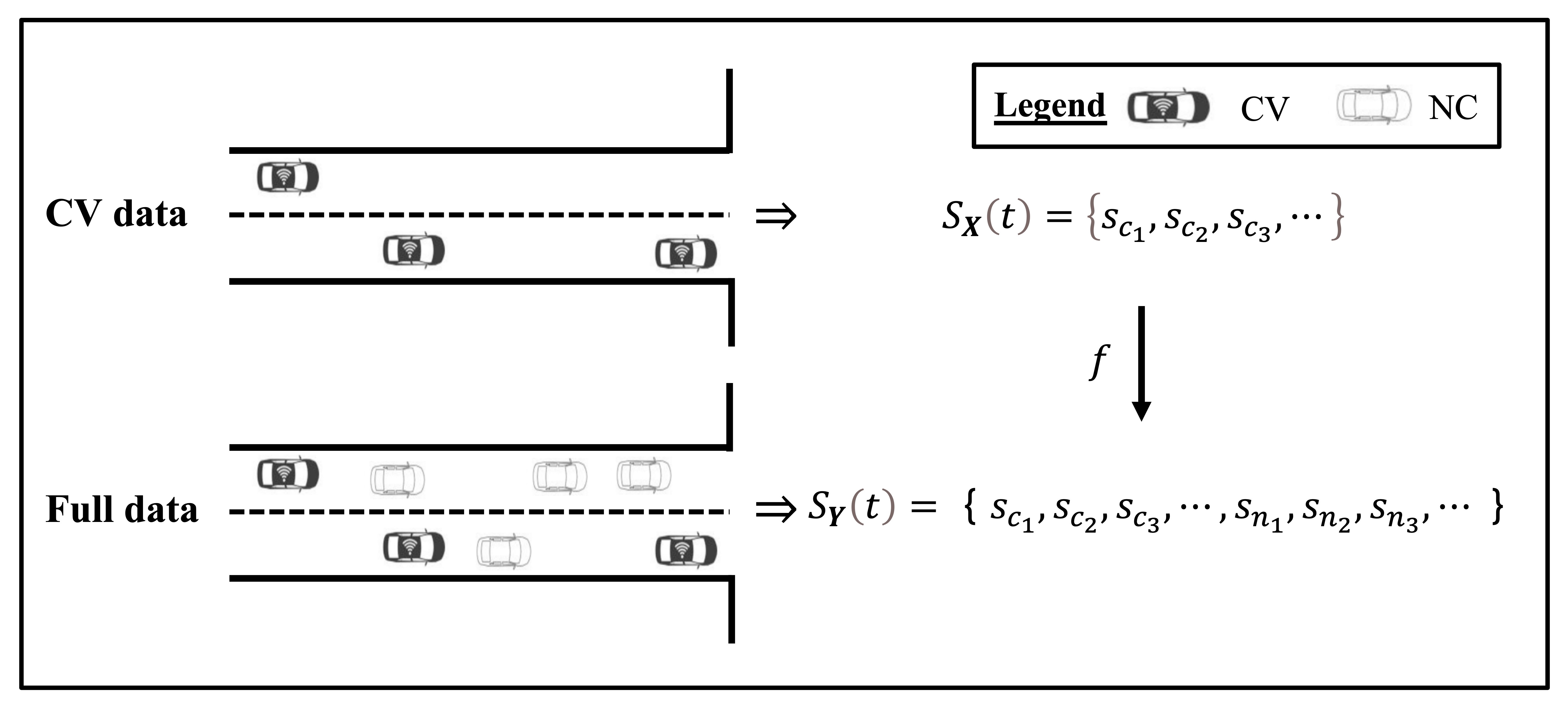}
      \caption{Illustration of vehicle state estimation.}
      \label{fig:ps}
\end{figure}

The goal of this paper is to estimate the locations and speeds of NCs based on the accessible CV data, as illustrated in Figure \ref{fig:ps}. Mathematically, it can be represented as: 
\begin{align}
    S_{Y}(t)=f[S_{X}(t),\cdots,S_{X}(t-k+1)], 
\end{align}
\noindent where $k$ is the number of past time steps considered; $f(\cdot)$ represents the mapping function, which will be learned.

\section{METHOD}
This section introduces the detailed method, including road cell occupancy (RCO) for data representation, CVVLSNet design, and loss functions used in training. 

\subsection{Road Cell Occupancy}

\begin{figure}[thpb]
      \centering
      \includegraphics[width=0.85\linewidth]{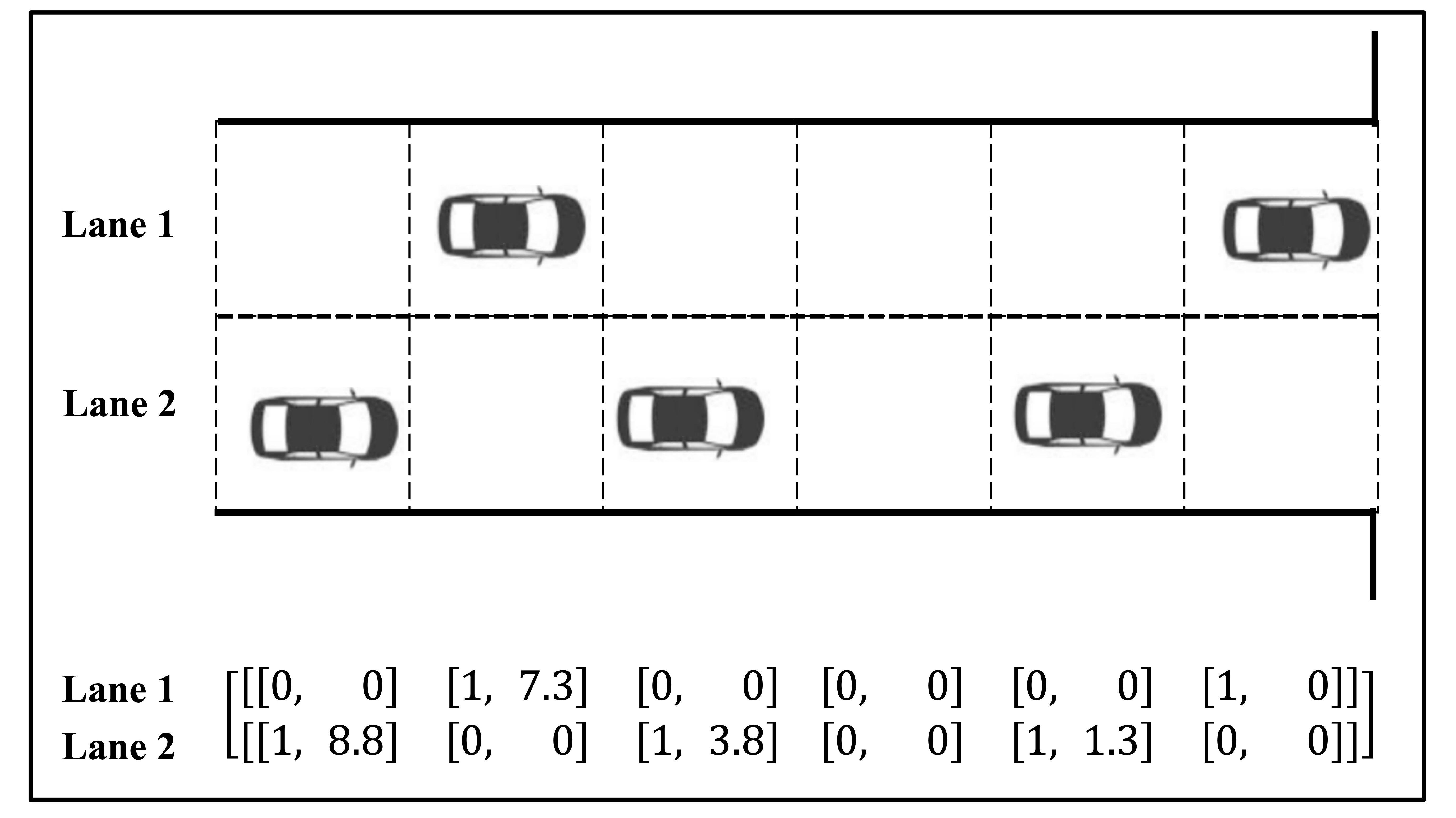}
      \caption{Illustration of RCO.}
      \label{fig:md1}
\end{figure}

According to the problem statement, the number of vehicles on the road is uncertain. Therefore, the lengths of the input and output can vary. Given that neural networks typically require fixed-size inputs and outputs, a road cell occupancy (RCO), as depicted in Figure \ref{fig:md1}, is proposed to transfer the varied number of vehicles to fixed-size representations. 

The key idea is to divide a lane into many cells with a predefined interval. The state of each cell is denoted as $[occ, spd]$, where $occ = 1$ represents that there is a vehicle in the cell, 0 otherwise; $spd$ is equal to the speed of the vehicle in the cell when $spd = 1$, -1 otherwise. Taking the cell length as 1 m gives $d$ cells in each lane. With RCO, the input and output information are converted to fixed size matrices, as shown below 
\begin{align}
S_{X}(t) \overset{\text{RCO}}{\longrightarrow}X(t) \in \mathbb{R}^{l\times d \times 2},  \nonumber \\
    S_{Y}(t) \overset{\text{RCO}}{\longrightarrow}Y(t) \in \mathbb{R}^{l\times d \times 2}.
\end{align}

Taking temporal information into account, the inputs at different past time steps are concatenated at the last dimension. Thus, the problem can be rewritten as
\begin{align}
    Y(t)=f\{concat[X(t),\cdots,X(t-k+1)]\}, 
\end{align}
\noindent where $concat(\cdot)$ represents the concatenation operation.

\subsection{CVVLSNet}

\begin{figure}[thpb]
      \centering
      \includegraphics[width=0.90\linewidth]{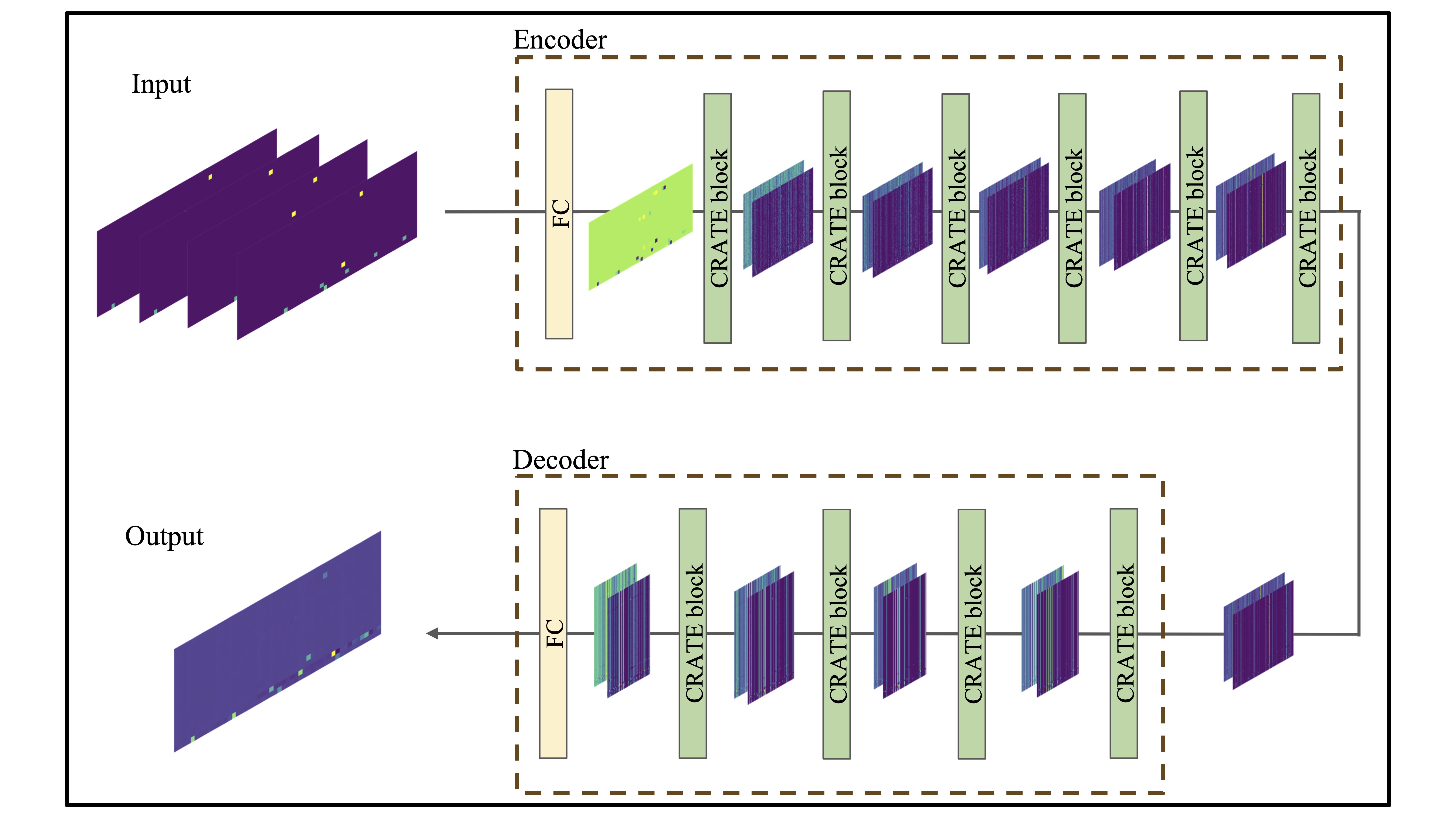}
      \caption{Architecture of CVVLSNet.}
      \label{fig:md2}
\end{figure}

A CV-based vehicle location and speed network, i.e., CVVLSNet, is proposed for simultaneously estimating vehicle locations and speeds based only on the partial CV trajectory data. An encoder-decoder architecture is adopted, as shown in Figure \ref{fig:md2}. The encoder consists of a fully connected (FC) layer and six CRATE (Coding-RAte TransformEr) \cite{32} blocks. The decoder comprises four CRATE blocks and a final FC layer for prediction. 

The FC layer in the encoder is to connect the input matrix and CRATE blocks, aiming at extracting features and fusing inputs from different time steps. This FC layer can also be used to reduce the input size and reshape features to a predefined size, thereby matching the input size of CRATE blocks. 

\begin{figure}[thpb]
      \centering
      \includegraphics[width=0.85\linewidth]{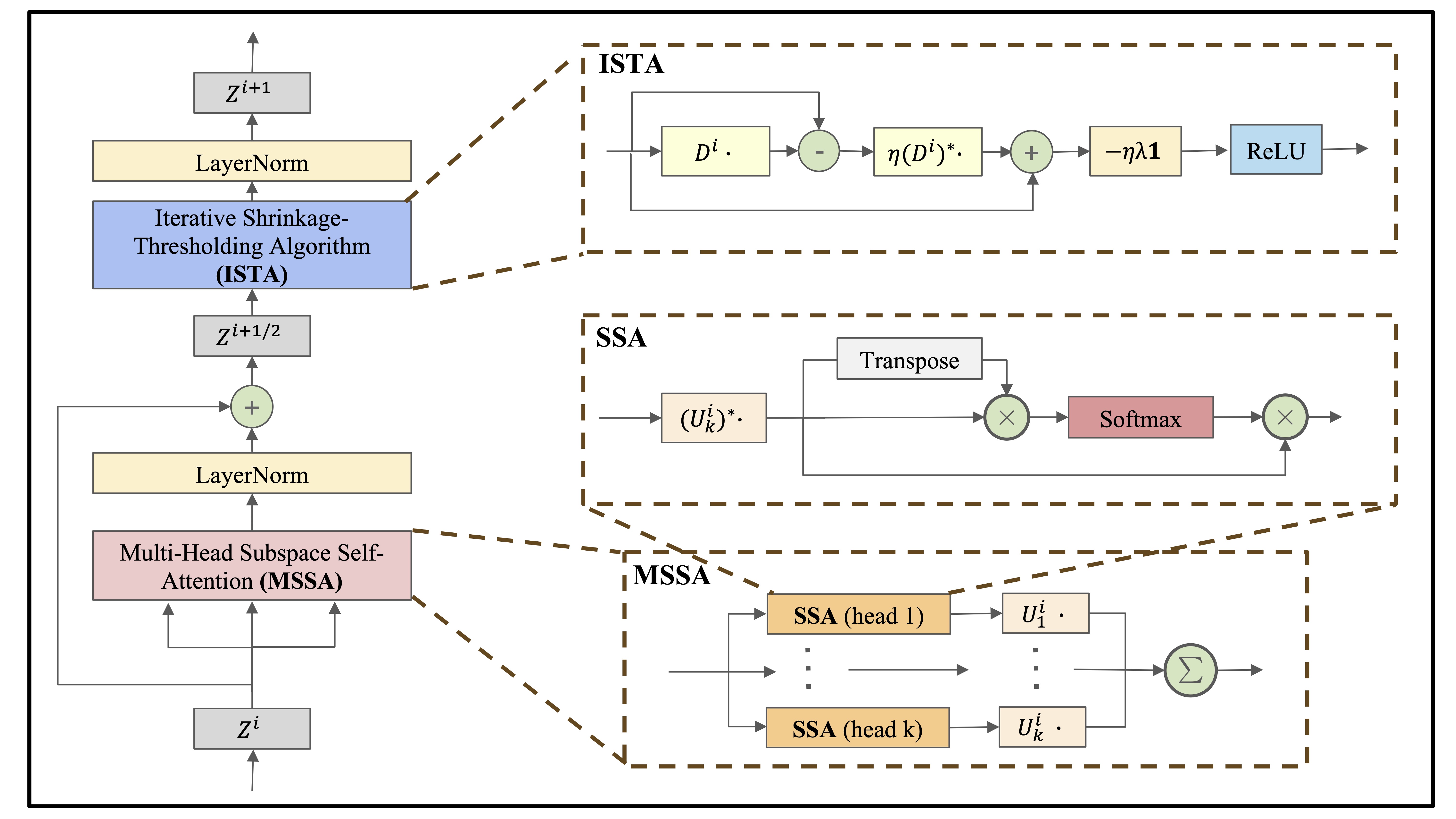}
      \caption{Architecture of CRATE blocks.}
      \label{fig:md3}
\end{figure}

A CRATE block, as shown in Figure \ref{fig:md3}, is proposed to maximize the information gain through coding rate reduction and improve the sparsity of the learned features. CRATE features a white-box alternative to Transformer \cite{33}. 

For a specific feature map, $Z \in \mathbb{R}^{h \times N}$, which is the coding rate, representing the ratio of useful information, can be defined as
\begin{align}
    R(Z)=\frac{1}{2}log\left[det\left( I+\frac{h}{N\epsilon^2}ZZ^* \right) \right],
\end{align}
\noindent where $det(\cdot)$ represents the determinant of a matrix; $I$ is the identity matrix; $\epsilon$ is the quantization precision that satisfies $\epsilon > 0$; $*$ is transpose operation.

Assume that $Z$ is composed of a series of Gaussian distributions and denote the orthogonal basis of Gaussian space as $U_{[K]}=(U_k)^{K}_{k=1}$ , where $U_k\in \mathbb{R}^{h\times w}$. Thus, the coding rate reduction under this assumption, $\Delta R(Z;U_{[K]})$, is given by 
\begin{align}
    \Delta R(Z;U_{[K]})=R(Z)- R(Z;U_{[K]}),
\end{align}
\noindent where $R(Z;U_{[K]})$ represents the coding rate when the specific orthogonal basis is known.

Therefore, the optimization objective of CRATE can be written as

$\ \ \ \ \ \ \ \ \ \ \underset{U_{[K]}}{max} \mathbb{E}[\Delta R(Z) - \lambda \|Z\|_0]$
\begin{align}
= \underset{U_{[K]}}{max} \mathbb{E}[R(Z) - \sum_{k=1}^K R(U^*_k Z) - \lambda \|Z\|_0],
\end{align}
\noindent where $\lambda \|Z\|_0$ is the regularization term which represents the sparsification of $Z$; $\|\cdot\|_0$ represents the zero-norm; $\lambda$ is regularization factor.

Two core blocks are introduced to solve the optimization problem: Multi-Head Subspace Self-Attention (MSSA) block and Iterative Shrinkage-Thresholding Algorithm (ISTA) block. They are detailed below, respectively.

The MSSA block is used to minimize $\sum^K_{k=1}R(U^*_kZ)$ in Eq. (7). Denote the input and the orthogonal basis of Gaussian space of $i^{th}$ CRATE block as $Z^i$ and $U^i_{[K]}$ . In this module, the feature vector $Z^i$ is iteratively updated through the followings: 
\begin{align}
    Z^{i+\frac{1}{2}}&=Z^i-\kappa \bigtriangledown R(Z^i;U^i_{[K]}) \nonumber \\
    &\approx \left(1-\frac{w}{N\epsilon^2}\cdot \kappa \right)Z^i+\frac{w}{N\epsilon^2}\cdot \kappa\cdot MSSA(Z^i|U^i_{[K]}),
\end{align}
\begin{align}
    MSSA(Z^i|U^i_{[K]})=\frac{w}{N\epsilon^2}\cdot[U^i_i,\cdots,U^i_k]\begin{bmatrix}
SSA(Z^i|U^i_1) \\
\vdots \\
SSA(Z^i|U^i_k)
\end{bmatrix},
\end{align}
\begin{small}
\begin{align}
SSA(Z^i|U^i_k)=\left((U^i_k)^*Z^i\right)softmax[\left((U^i_k)^*Z^i\right)^*\left((U^i_k)^*Z^i\right)],
\end{align}
\end{small}

\noindent where $\kappa$ is the learning rate; $k\in [K], Z^{i+\frac{1}{2}}$ is the output of MSSA block as an intermediate feature vector between $Z^i$ and $Z^{i+1}$. The SSA operator in Eq. (10) is similar with the attention operator in Transformer, as all value, key and query in Transformer are set to $U_k$ . 

The ISTA block is used to maximize $[R(Z)- \lambda \|Z\|_0]$ in Eq. (7). To avoid the low scalability of $R(Z)$’s gradient, a complete orthogonal dictionary, $D\in \mathbb{R}^{h\times h}$ , is introduced based on the fact that 
\begin{align}
    R(Z^{i+1})\approx R(DZ^{i+1})=R(Z^{i+\frac{1}{2}}) \ \ \ s.t.\ DD^*\approx I.
\end{align}

Thus, an approximate optimization program is derived as:
\begin{align}
    Z^{i+1}=arg\underset{Z}{min} \|Z\|_0 \ \ \  s.t. \  Z^{i+\frac{1}{2}}=D^iZ^{i+1}.
\end{align}
\noindent where $D^i$ is the complete orthogonal dictionary of $i^{th}$ CRATE block.

The least absolute shrinkage and selection operator (LASSO) method \cite{34} is used to solve it with a non-negative constraint. The problem is converted to the following minimization:
\begin{align}
    Z^{i+1}=arg \underset{Z\geq0}{min}\left[\lambda\|Z\|_1+\frac{1}{2}\|Z^{i+\frac{1}{2}}-D^iZ^{i+1}\|^2_F \right],
\end{align}
\noindent where $\|\cdot\|_F$ represents the F-norm. An unroll proximal gradient descent step is used to incrementally optimize the variables, i.e., 
\begin{small}
\begin{align}
    Z^{i+1}&=ISTA(Z^{i+\frac{1}{2}}|D^i) \nonumber \\
    &=ReLU(Z^{i+\frac{1}{2}}-\eta (D^i)^*\left(D^iZ^{i+\frac{1}{2}}-Z^{i+\frac{1}{2}}\right)-\eta \lambda 1),
\end{align}
\end{small}

\noindent where $ReLU(\cdot)$ is the ReLU activation function; $\eta$ is the step size for the approximate gradient ascent; $\lambda$ is the sparsification regularization term; $1$ is an all-1 adaptive-shape tensor. 

\subsection{Loss functions}

The mean square error (MSE) loss is used as the major loss for training CVVLSNet, as shown below: 
\begin{align}
    L_M=\frac{1}{m}\sum^m_{i=1}(\hat{y_i}-y_i)^2,
\end{align}
\noindent where $\hat{y_i}$ and $y_i$ represent the $i^{th}$ estimated and ground truth vehicle state.

Meanwhile, for each occupied cell, its preceding and following several cells must be empty, due to physical size constraints. This number of empty cells before or after this occupied cell, $d_e$ , can be determined by the average effective vehicle length. Let $NLS(\cdot)$ represent the nearby vehicles’ location sum, which is defined as 
\begin{align}
    NLS(i,l_0)=\sum^{i+d_e}_{j=i-d_e}occ_{j,l_0},
\end{align}
\noindent where the $i^{th}$ cell in lane $l_0$ is occupied and $occ_{j,l_0}$ represents the occupancy condition of the $j^{th}$ cell in lane $l_0$. On this basis, a safety penalty term, $L_p$ , is introduced as 
\begin{align}
    L_p=\sum_l\sum_{i\in \Omega}(NLS(i,l)-1)^2,
\end{align}
\noindent where $\Omega$ represents all occupied cells.

Therefore, the total loss is given by
\begin{align}
    L = L_M+\mu L_p,
\end{align}
\noindent where $\mu$ is the weighting factor.

\section{EXPERIMENT}

This section conducts comprehensive experiments to demonstrate the effectiveness of the proposed method. Datasets, training details, metrics, results, sensitivity analysis, and feature analysis are introduced in order. 

\subsection{Datasets}

The Simulation of Urban Mobility (SUMO) platform is used to generate training and testing data. Consider a single signalized lane with a length of 1 km. The free flow speed was set to 50 km/h. The cycle length and amber period were set to 60 s and 3 s, respectively. Different red periods, i.e., 15 s, 30 s, and 45 s, were used to generate various signal plans. 

The vehicle length, minimum gap between two vehicles, car-following model, and driver's desired time headway were set to 5 m, 2.5 m, intelligent driver model (IDM), and 1 s. The departure mode of vehicles was set to “max”, i.e., vehicles would be generated with the free flow speed, while the speed could be adjusted to ensure a safe distance to the preceding vehicle. Each vehicle had the same probability (i.e., CV penetration rate) to be a CV. The vehicle arrival pattern followed a Poisson distribution. The volume-to-capacity (V/C) ratios were set to 0.3, 0.6, and 0.9. Thus, there were 9 combinations of signal plans and V/C ratios in total, forming 9 simulation scenarios. In all simulation experiments, the simulation resolution was set as 0.1 s, which means that all vehicle states would be updated 10 times per second. Each simulation scenario was run for 60 cycles (3600 s). The detailed trajectory data were recorded for training and testing. 

The first 10 cycles in each simulation scenario were used for warming up. After the warming-up period, 45-cycle data in each simulation scenario were used for training the network; the remaining 5-cycle data in each simulation scenario were used for testing. In total, the training and testing sets include 405-cycle data and 45-cycle data, respectively. The vehicle locations and speeds were estimated at every second during the testing phase. 

\subsection{Training details}

The initial learning rate, weight decay, momentum, and batch size were set to $4\times10^{-4}$, 0.1, 0.9, and 256, respectively. Each model was trained for 20 epochs. AdamW optimizer was used. 

\subsection{Metrics}
In order to quantitatively evaluate the performance, each lane is divided into segments by CVs, the estimates are matched with the ground truths within each segment. If the distance between two matched vehicles is less than a predefined threshold, this sample is considered a true positive (TP). If the distance between two matched vehicles is greater than the predefined threshold or the estimated vehicle is not in ground truth, this sample is considered a false positive (FP). Those existing in the ground truths but are not estimated are considered false negative (FN). Following these definitions, precision, recall, and F1 score can be calculated as 
\begin{align}
    Precision&=\frac{TP}{TP+FP}, \\
    Recall&=\frac{TP}{TP+FN}, \\
    F1\ score &= 2*\frac{Presicion*Recall}{Precision + Recall}.
\end{align}

For speed estimation, root mean square error (RMSE) is used to measure the performance, which is calculated as:
\begin{align}
    RMSE=\sqrt{\frac{\sum(v_{tp}-v_{gt})^2}{N_{tp}}},
\end{align}
\noindent where $N_{tp}$ is the number of true positive samples; $v_{tp}$ and $v_{gt}$ represent the estimated and truthful speeds of a true positive sample.

\subsection{Results}

The quantitative results of vehicle location and speed estimation using the proposed CVVLSNet and the existing EVLS algorithm \cite{1} are presented in Table \ref{tab:rs-total}. ↑ represents performance improvement. It was found that the proposed CVVLSNet significantly outperformed the ELVS algorithm under various CV penetration rates and V/C ratios. 

\begin{table*}[!th]
    \centering
    \caption{RESULTS OF LOCATION AND SPEED ESTIMATIONS FROM THE EVLS ALGORITHM \cite{1} AND CVVLSNET.}
    \begin{tabular}{cccccccccccccc}
        \toprule
        \multicolumn{2}{c}{Parameters} & \multicolumn{9}{c}{Location Estimation} & \multicolumn{3}{c}{Speed Estimation}  \\
        \cmidrule{3-14}
        \multirow{2}{*}{\makecell{CV\\ Penetration Rate}} & \multirow{2}{*}{V/C Ratio} & \multicolumn{3}{c}{Precision (\%)} & \multicolumn{3}{c}{Recall (\%)} & \multicolumn{3}{c}{F1 Score (\%)} & \multicolumn{3}{c}{RMSE (m/s)} \\ 
        \cmidrule{3-14}
        & & EVLS & OURS & ↑ & EVLS & OURS & ↑ & EVLS & OURS & ↑ & EVLS & OURS & ↑ (\%) \\ 
        \midrule
        \multirow{4}{*}{0.1} & 0.3 & 3.8 & 40.3 & 36.5 & 1.5 & 10.0 & 8.5 & 2.0 & 16.0 & 14.0 & 4.4 & 4.0 & 9.1 \\
        & 0.6 & 2.9 & 27.2 & 24.3 & 1.2 & 6.6 & 5.4 & 1.6 & 10.7 & 9.1 & 3.8 & 3.6 & 5.3 \\
        & 0.9 & 16.7 & 30.5 & 13.8 & 10.4 & 11.3 & 0.9 & 12.8 & 16.5 & 3.7 & 5.1 & 2.5 & 51.0 \\
        & Mean & 7.8 & 32.7 & 24.9 & 4.4 & 9.3 & 4.9 & 5.5 & 14.4 & 8.9 & 4.4 & 3.4 & 21.8 \\ 
        \midrule
        \multirow{4}{*}{0.4} & 0.3 & 5.2 & 79.5 & 74.3 & 9.5 & 51.5 & 42.0 & 6.3 & 62.5 & 56.2 & 3.9 & 3.8 & 2.6 \\
        & 0.6 & 10.1 & 76.1 & 66.0 & 19.6 & 50.4 & 30.8 & 12.9 & 60.6 & 47.7 & 3.5 & 3.3 & 5.7 \\
        & 0.9 & 21.6 & 82.3 & 60.7 & 40.9 & 58.6 & 17.7 & 27.3 & 68.4 & 41.1 & 4.1 & 2.4 & 41.5 \\
        & Mean & 12.3 & 79.3 & 67.0 & 23.3 & 53.5 & 30.2 & 15.5 & 63.8 & 48.3 & 3.8 & 3.2 & 16.6 \\ 
        \midrule
        \multirow{4}{*}{0.7} & 0.3 & 2.2 & 95.4 & 93.2 & 14.6 & 74.5 & 59.9 & 3.7 & 83.7 & 80.0 & 3.5 & 3.4 & 2.9 \\
        & 0.6 & 4.7 & 94.2 & 89.5 & 26.9 & 76.9 & 50.0 & 7.9 & 84.7 & 76.8 & 3.1 & 3.0 & 3.2 \\
        & 0.9 & 9.9 & 95.3 & 85.4 & 54.1 & 79.8 & 25.7 & 16.5 & 86.9 & 70.4 & 4.2 & 2.3 & 45.2 \\
        & Mean & 5.6 & 95.0 & 89.4 & 31.9 & 77.1 & 45.2 & 9.4 & 85.1 & 75.7 & 3.6 & 2.9 & 17.1 \\ 
        \bottomrule
    \end{tabular}
    \label{tab:rs-total}
\end{table*}

Specifically, with a CV penetration rate of 0.1, the average improvements in location estimation across various V/C ratios on precision, recall, and F1 score were found to be 24.9\%, 4.9\%, and 8.9\%, respectively. This clearly demonstrates the effectiveness of the proposed CVVLSNet in CV-based location estimation. By increasing CV penetration rate to 0.4, the average improvements were increased to 67\%, 30.2\%, and 48.3\%. Further increasing CV penetration rate to 0.7, the average improvements were up to 89.4\%, 45.2\%, and 75.7\%. All these numbers indicate the huge advantages of the CVVLSNet in CV-based location estimation. 

In speed estimation, the average improvements across diverse V/C ratios were in the range of 16.6\% to 21.8\%, forming a relatively stable pattern. This further affirms the effectiveness and superiority of the proposed CVVLSNet in CV-based speed estimation. 

Overall, the proposed CVVLSNet exhibited excellent performance in both CV-based location and speed estimation and has great potential to be used in many downstream applications, such as adaptive signal control. 

\subsection{Sensitivity analysis}
A series of sensitivity analyses were conducted to observe the performance changes with various numbers of past time steps considered, $k$. 

When $k$ varies from 1 to 6, the location and speed estimation results are reported in Table \ref{tab:f1} and Table \ref{tab:rmse}. It indicates that (1) the location estimation performance can be improved as $k$ increases from 1 to 4 while further increasing $k$ does not observe consistent improvements; (2) the performance of speed estimation is relatively stable and does not exhibit a clear pattern with respect to $k$; and (3) $k=4$ provides the best speed estimation results. Therefore, $k$ was set to 4 in this paper. 

\begin{table}[!h]
    \centering
    \caption{RESULTS OF LOCATION ESTIMATION USING CVVLSNET WITH DIFFERENT k.}
    \begin{tabular}{cccc}
        \toprule
        \multirow{3}{*}{k} & \multicolumn{3}{c}{CV Penetration Rate} \\
        \cmidrule{2-4}
        & 0.1 & 0.4 & 0.7 \\ 
        \cmidrule{2-4}
        & F1 score (\%) & F1 score (\%) & F1 score (\%) \\ 
        \midrule
        1 & 5.6 & 37.0 & 69.6 \\ 
        2 & 9.3 & 53.4 & 81.5 \\ 
        3 & 10.8 & 57.4 & 82.0 \\ 
        4 & 14.4 & 63.8 & 85.1 \\ 
        5 & 13.5 & 62.4 & 82.9 \\ 
        6 & 13.7 & 64.3 & 85.3 \\ 
        \bottomrule
    \end{tabular}
    \label{tab:f1}
\end{table}

\begin{table}[!ht]
    \centering
    \caption{RESULTS OF SPEED ESTIMATION USING CVVLSNET WITH DIFFERENT k.}
    \begin{tabular}{cccc}
        \toprule
        \multirow{3}{*}{k} & \multicolumn{3}{c}{CV Penetration Rate} \\
        \cmidrule{2-4}
        & 0.1 & 0.4 & 0.7 \\ 
        \cmidrule{2-4}
        & RMSE (m/s) & RMSE (m/s) & RMSE (m/s) \\ 
        \midrule
        1 & 3.6 & 3.4 & 3.5 \\ 
        2 & 3.6 & 3.4 & 3.4 \\ 
        3 & 4.1 & 3.8 & 3.7 \\ 
        4 & 3.4 & 3.2 & 2.9 \\ 
        5 & 3.7 & 3.5 & 3.5 \\ 
        6 & 3.9 & 3.4 & 3.3 \\ 
        \bottomrule
    \end{tabular}
    \label{tab:rmse}
\end{table}

\subsection{Feature analysis}

\begin{figure}[thpb]
      \centering
      \includegraphics[width=0.85\linewidth]{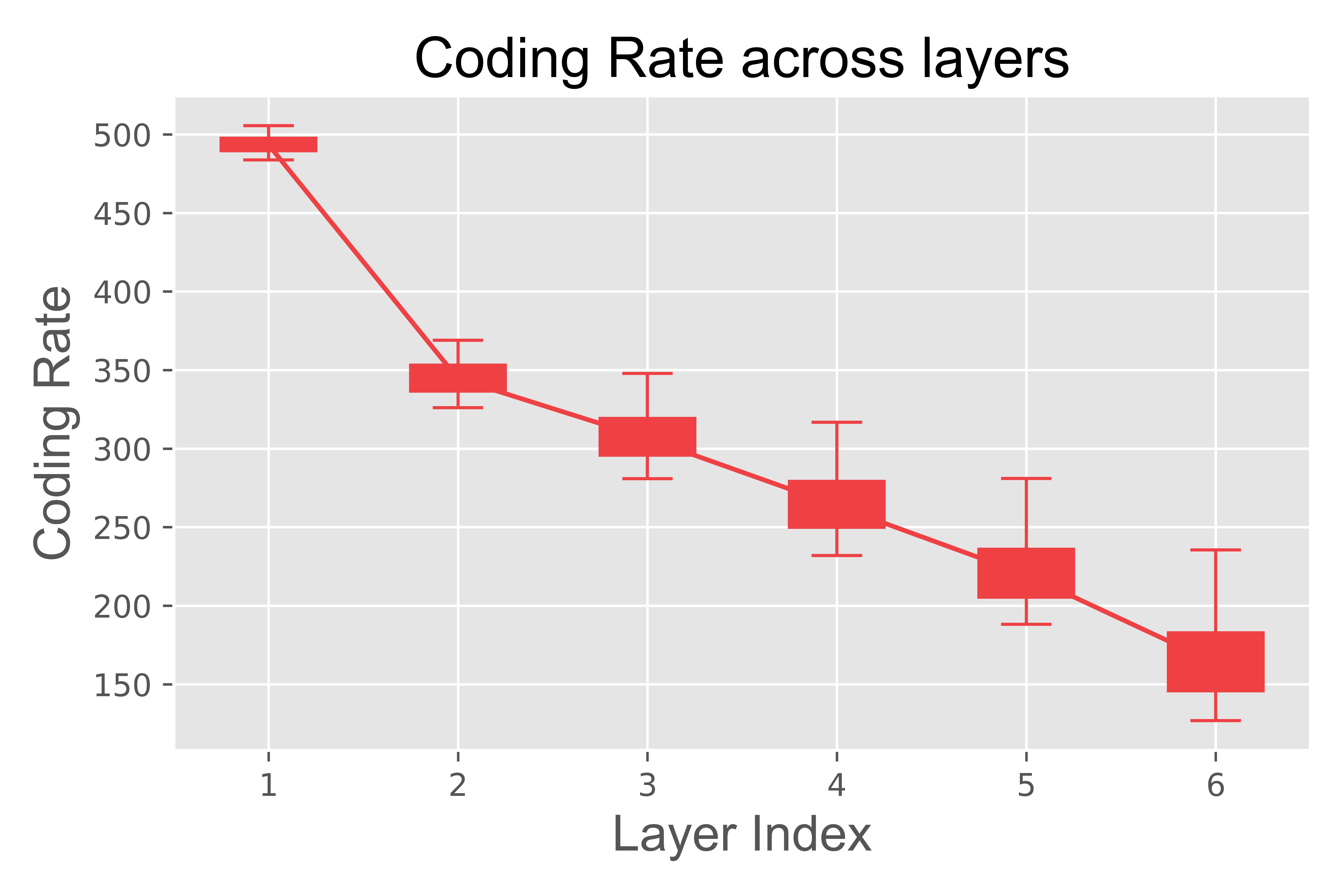}
      \caption{Coding rates of different layers in the encoder of CVVLSNet.}
      \label{fig:rs}
\end{figure}

This part is to count the features in CVVLSNet to demonstrate the working mechanism. The coding rates of each layer in the encoder of CVVLSNet are computed and shown in Figure \ref{fig:rs}. As expected, the coding rates steadily decrease as the increase of network depth, indicating that the representations become more and more compressed. This validates the working mechanism in CVVLSNet and also reflects the mathematical interpretability of the model. 

\section{CONCLUSIONS}

This paper proposes a CV-based vehicle location and speed estimation network, CVVLSNet, to simultaneously estimate vehicle locations and speeds, by solely using accessible CV trajectory data. The proposed method does not require any fixed detectors on the roads and market penetration rate, and thus is highly flexible. Comprehensive experiment results show that CVVLSNet exhibits superior performance as compared with the existing method. It is noted that although this paper conducted the experiments with a single lane, the method is applicable for an intersection by simply applying the method to each lane in the intersection. Taking the estimated location and speed as inputs, many downstream applications can be developed, such as estimations of vehicle arrival patterns and travel time, as well as adaptive signal control. 

\section{LIMITATION AND FUTURE WORK}
This paper has the following limitations. As this study only takes the CV states in the target lane as input to estimate full vehicle states, the correlations between adjacent lanes and intersections are not considered. Due to regular stops of vehicles during red periods in a network, the CV trajectories in upstream lanes may provide extra information for estimating NCs in the target lane. 

Future works include addressing the aforementioned issues and extending the presented method to conduct vehicle state estimation in a network. 




\end{document}